\documentclass{article}

\usepackage{microtype}
\usepackage{graphicx}
\usepackage{subfigure}
\usepackage{booktabs} % for professional tables

\usepackage{hyperref}

\usepackage[accepted]{icml2023}

\usepackage{amsmath}
\usepackage{amssymb}
\usepackage{mathtools}
\usepackage{amsthm}

\usepackage[capitalize,noabbrev]{cleveref}

\theoremstyle{plain}

\theoremstyle{definition}

\theoremstyle{remark}

\usepackage[textsize=tiny]{todonotes}

\begin{document}

\twocolumn[
\icmltitle{AutoML-GPT: Large Language Model for AutoML}




\begin{icmlauthorlist}
\icmlauthor{Yun-Da Tsai}{NTU}
\icmlauthor{Yu-Che Tsai}{NTU}
\icmlauthor{Bo-Wei Huang}{NTU}
\icmlauthor{Chun-Pai Yang}{NTU}
\icmlauthor{Shou-De Lin}{NTU}
\end{icmlauthorlist}

\icmlaffiliation{NTU}{Department of Computer Science and Information Engineering, National Taiwan University, Taipei, Taiwan}

\icmlcorrespondingauthor{Yun-Da Tsai}{f08946007@csie.ntu.edu.tw}

\icmlkeywords{Machine Learning, ICML}

\vskip 0.3in
]



\printAffiliationsAndNotice{}  % leave blank if no need to mention equal contribution

\begin{abstract}
With the emerging trend of GPT models, we have established a framework called AutoML-GPT that integrates a comprehensive set of tools and libraries. This framework grants users access to a wide range of data preprocessing techniques, feature engineering methods, and model selection algorithms. Through a conversational interface, users can specify their requirements, constraints, and evaluation metrics.
Throughout the process, AutoML-GPT employs advanced techniques for hyperparameter optimization and model selection, ensuring that the resulting model achieves optimal performance. The system effectively manages the complexity of the machine learning pipeline, guiding users towards the best choices without requiring deep domain knowledge.
Through our experimental results on diverse datasets, we have demonstrated that AutoML-GPT significantly reduces the time and effort required for machine learning tasks. Its ability to leverage the vast knowledge encoded in large language models enables it to provide valuable insights, identify potential pitfalls, and suggest effective solutions to common challenges faced during model training.
\end{abstract}

\section{Introduction}
% intro: automl
Automated Machine Learning (AutoML) has gained significant attention in recent years as a powerful technique for automating various stages of the machine learning workflow. It aims to simplify the model development process by automatically searching, selecting, and optimizing machine learning models without requiring extensive manual intervention. AutoML has the potential to democratize machine learning and make it accessible to a broader audience, including non-experts and domain specialists.
AutoML encompasses tasks such as data pre-processing, feature engineering, model selection, and hyperparameter tuning. These tasks require expertise, time, and computational resources. To address these challenges, researchers have proposed various approaches and frameworks for AutoML, such as Auto-sklearn~\cite{autosklearn}, Auto-Keras~\cite{autokeras}, and AutoGluon~\cite{autoGluon}. These approaches aim to automate the process of model selection and configuration, making it easier for practitioners to build accurate and efficient machine learning models.

% intro: llm
Large language models, such as OpenAI's GPT-3~\cite{GPT3} and Google's PaLM~\cite{chowdhery2022palm}, have emerged as powerful tools in natural language processing and understanding. These models have been extensively trained on vast amounts of textual data and have demonstrated remarkable capabilities in language understanding, text generation, sentiment analysis, and other language-related tasks. Large language models excel at capturing complex patterns, understanding context, and generating coherent responses.
The strength of large language models lies in their ability to comprehend and process unstructured textual data effectively. They learn rich semantic representations of language, enabling them to understand the nuances and subtleties present in text. By leveraging pre-trained language models, researchers have achieved significant advancements in various natural language processing tasks, including machine translation, question answering, and language generation.

% motivation: integrate llm into automl
While large language models have found success in specific applications, their comprehensive integration into the AutoML framework remains relatively unexplored. Existing research has primarily focused on utilizing language models for individual tasks within AutoML, such as data pre-processing and feature engineering. However, the potential of leveraging these models to automate the entire AutoML pipeline, from data preparation to hyperparameter optimization, has not been extensively studied. This lack of exploration serves as the motivation for this work.
Large language models (LLMs) exhibit exceptional strengths in AutoML from multiple perspectives. In terms of data understanding, LLMs can be used for data preprocessing, efficiently handling missing values, normalizing or scaling features, and detecting outliers. Additionally, large language models have the ability to conduct correlation analysis, discover causal relationships, and perform feature selection. This enables them to effectively identify and eliminate irrelevant or redundant features. Furthermore, these models contribute to the identification of potential models that are well-suited for a given dataset and task, providing valuable guidance in the model selection phase.

% research problem
In this paper, we designed AutoML-GPT, a dual agent system built upon large language models. The agents in the system are capable of communication, planning, and using tools to complete complex machine learning tasks. In our experiments, AutoML-GPT demonstrated compatible performance compared to human experts on 11 tabular datasets chosen from recent Kaggle competitions, reflecting real modern-day ML applications.

\section{Methodology}
\begin{figure}[t!]
  \centering
  \includegraphics[width=1\linewidth]{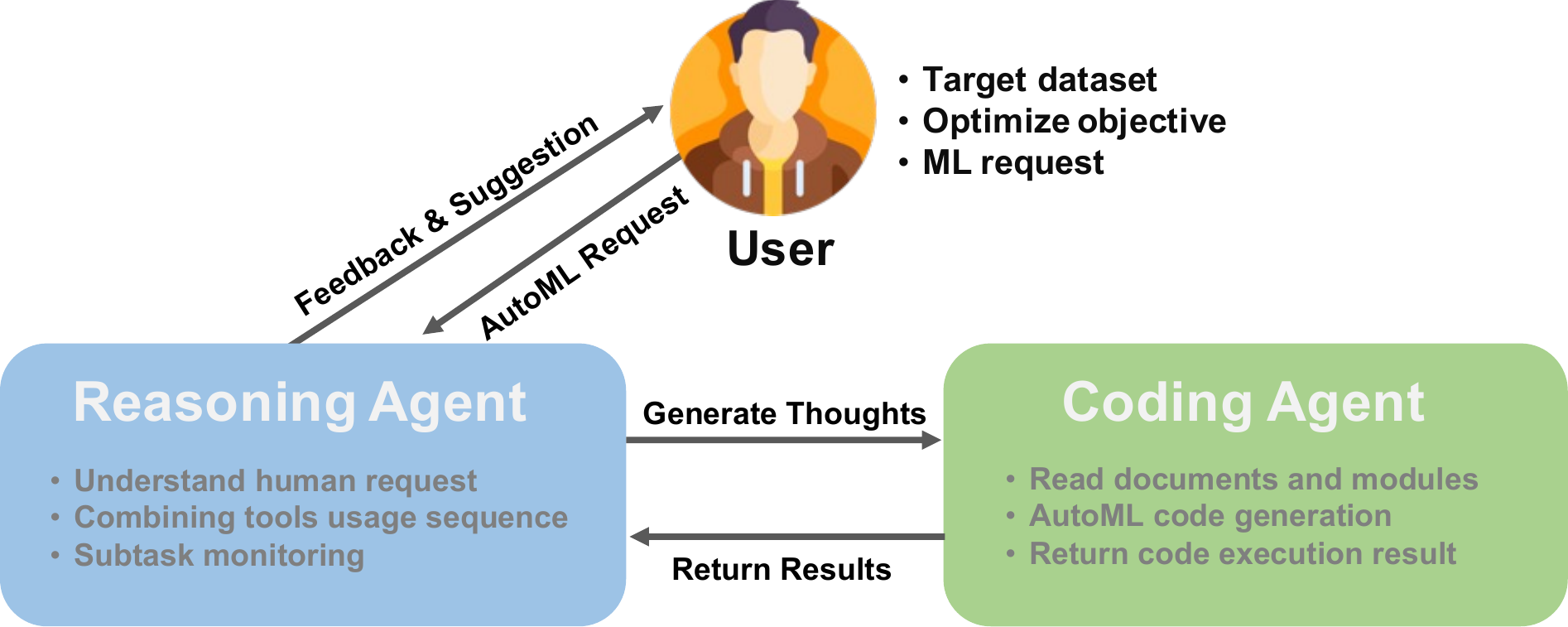}
  \caption{Pipeline of AutoML-GPT.}
  \label{fig:autoML_pipeline}
  \vspace{-10pt}
\end{figure}
To integrate a Large Language Model (LLM) into the AutoML framework, we propose a systematic methodology that involves two agents: the Reasoning agent and the Coding agent, as illustrated in Figure~\ref{fig:autoML_pipeline}.
Both agents are implemented using the ReAct~\cite{yao2022react} framework by langchain~\footnote{https://github.com/hwchase17/langchain}.

The Reasoning agent in the AutoML pipeline handles the task of understanding human requests and planning the sequence of tool usage. It utilizes the language comprehension capabilities of the LLM to accurately interpret complex requests and effectively plan the steps required for tasks like end-to-end training. This agent is responsible for combining different tools optimally, monitoring the pipeline's progress, and providing timely updates to the user.
On the other hand, the Coding agent is responsible for implementing the planned tasks. It acquires the necessary knowledge by reading documentation and modules, leveraging its understanding of programming languages and AutoML tools. The Coding agent generates ideas, formulates code, and executes it in a structured manner to carry out the specified actions within the AutoML pipeline. It plays a vital role in translating the reasoning and planning of the Reasoning agent into executable code.

The interaction between the Reasoning and Coding agents is iterative and collaborative. The Reasoning agent receives the execution output from the Coding agent and utilizes it to provide relevant and informative responses to the user. This enables the Reasoning agent to communicate the progress of the AutoML pipeline, respond to user queries, and deliver meaningful insights based on the executed tasks.
By employing this methodology with the Reasoning and Coding agents, the integration of LLM into the AutoML framework benefits from the reasoning and planning capabilities of the Reasoning agent, as well as the code generation and execution expertise of the Coding agent. This collaborative approach ensures the accurate interpretation of user requests, precise planning of tool usage, reliable code implementation, and effective communication with the user, facilitating a seamless and efficient AutoML pipeline.

\begin{figure}[t!]
  \centering
  \includegraphics[width=1\linewidth]{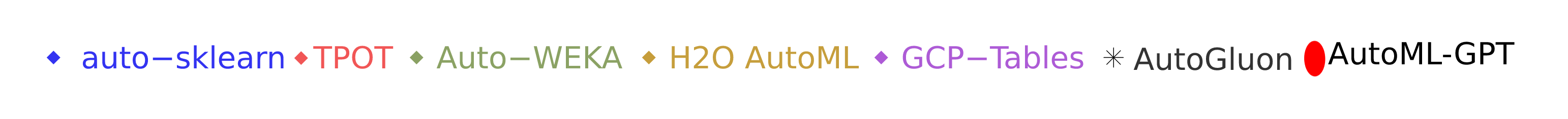}
  \includegraphics[width=1\linewidth]{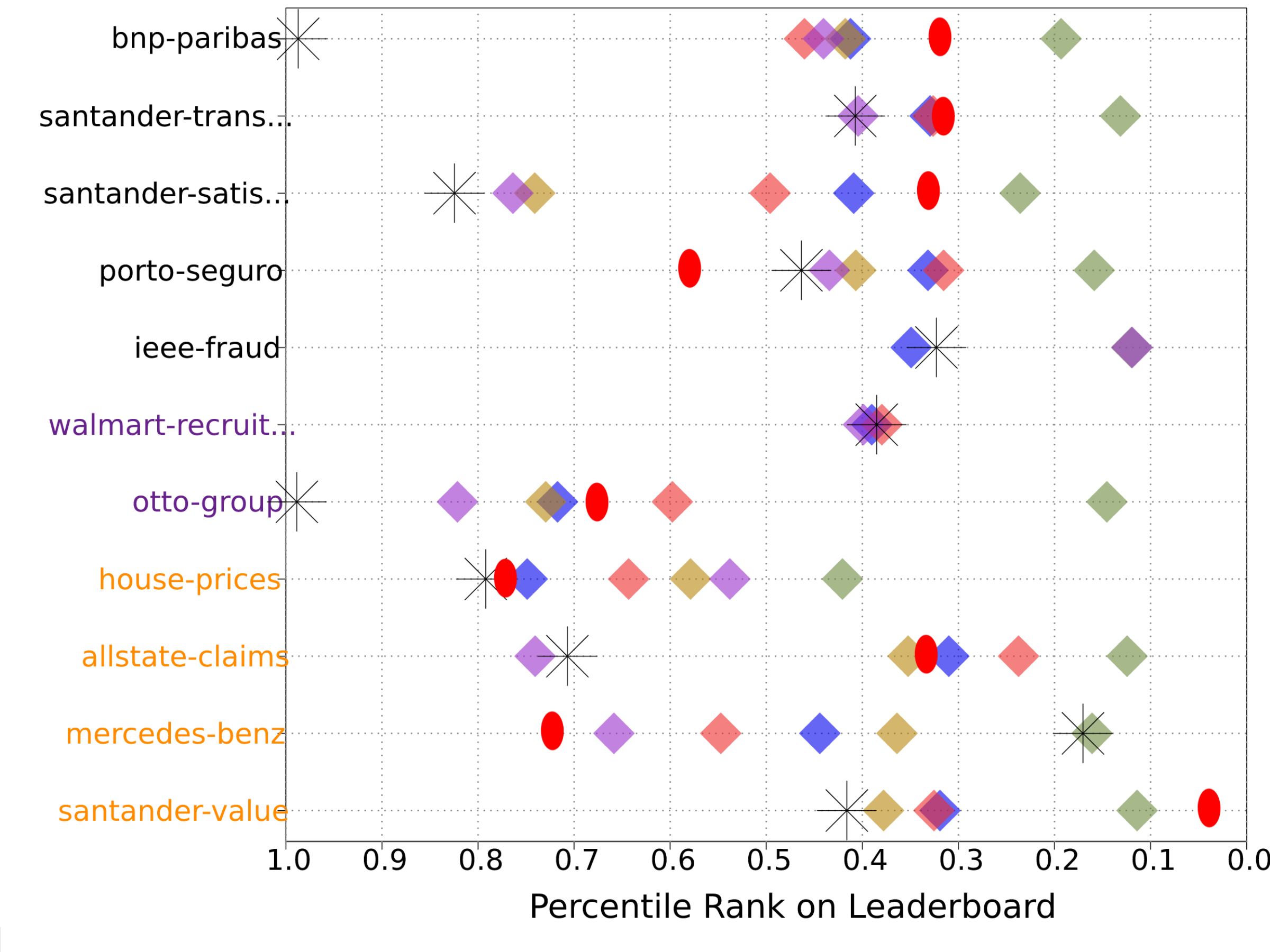}
  \vspace{-20pt}
  \caption{Summary of the performance in rank percentile on Leaderboard across 9 Kaggle competitions compared to other well known automl tools.}
  \label{fig:kaggle}
  \vspace{-10pt}
\end{figure}

\section{Experimental Results}
% dataset
For the experiment, we chose to utilize the widely used Kaggle benchmark, which holds significance in automl research literature. We selected nine tabular datasets from recent Kaggle competitions to represent contemporary applications. These competitions encompass a range of tasks, including regression and (binary/multiclass) classification. The competition organizers have tailored different metrics to evaluate the predictive performance based on the specific problem at hand.
% setup
Every dataset is processed by AutoML-GPT in a four instruction sequence though the LLM (1) Explore the dataset (2) Process the dataset (3) Select the model (4) Fine tune the parameters. In our experiment, we will use a single model without employing any ensemble techniques. The results, depicted in Figure~\ref{fig:kaggle}, represent the rank percentile on the Kaggle competition leaderboard. It is important to note that each result is a one-shot submission to Kaggle without any further fine-tuning after local development.
% baseline
In the experiment, we compared with 6 other renowned state-of-the-art automl frameworks: auto-sklearn, TPOT, Auto-WEKA, H2O AutoML, GCP-tables, AutoGluon. The experiment shown in Figure~\ref{fig:kaggle} limits the training time of each automl framework to 8 hours max.

\section{Discussion}
The experiment results in Figure~\ref{fig:kaggle} show AutoML-GPT with competitive performance. The key difference between AutoML-GPT and other automl framework is that most automl framework focus on tasks such as hyperparameter search and model ensemble techniques. The strength of the performance comes from extensive computation power. However, since we limit AutoML-GPT to single model result, the competitive performance comes from the expertise in machine learning domain knowledge. AutoML-GPT conducts great data exploration and understanding and thus create well processed datasets for model training. The performance will be further boost if we incorporate other automl frameworks as one of the tools into AutoML-GPT.

\section{Conclusion}
In this paper, we proposed the AutoML-GPT framework that uses LLM as machine learning expert to conduct auotml. We showed its competitive performance by comparing to other renowned automl frameworks and human competitiors on Kaggle benchmarks

\bibliography{example_paper}
\bibliographystyle{icml2023}

\end{document}